\documentclass{article}
\usepackage{spconf,amsmath,graphicx,epsfig}
\usepackage{verbatim}
\usepackage{xcolor}
\usepackage{amsfonts,todonotes}
\usepackage[algoruled,boxed,lined]{algorithm2e}
\usepackage{balance}
\makeatletter
\def\BState{\State\hskip-\ALG@thistlm}
\usepackage{booktabs}

\usepackage{multirow} %this is for multiple rows you don't need if you wont use it.
\usepackage{multicol}
\usepackage{rotating}

\usepackage{array}
\newcolumntype{L}[1]{>{\raggedright\arraybackslash}p{#1}}
\newcolumntype{C}[1]{>{\centering\arraybackslash}p{#1}}
\newcolumntype{R}[1]{>{\raggedleft\arraybackslash}p{#1}}
\title{Ensemble Network for Ranking Images Based on Visual Appeal}
\name{Sachin Singh$^1$ \hfill Victor Sanchez$^2$ \hfill Tanaya Guha$^2$}
\address{$^1$IIT Kanpur, India $\,\,\,\,$ $^2$University of Warwick, UK}
\begin{document}
\ninept
\maketitle
\begin{abstract}
We propose a computational framework for ranking images (group photos in particular) taken at the same event within a short time span. The ranking is expected to correspond with human perception of overall appeal of the images. We hypothesize and provide evidence through subjective analysis that the factors that appeal to humans are its emotional content, aesthetics and image quality. We propose a network which is an ensemble of three information channels, each predicting a score corresponding to one of the three visual appeal factors. For group emotion estimation, we propose a convolutional neural network (CNN) based architecture for predicting group emotion from images. This new architecture enforces the network to put emphasis on the important regions in the images, and achieves comparable results to the state-of-the-art. Next, we develop a network for the image ranking task that combines group emotion, aesthetics and image quality scores. Owing to the unavailability of suitable databases, we created a new database of manually annotated group photos taken during various social events. We present experimental results on this database and other benchmark databases whenever available. Overall, our experiments show that the proposed framework can reliably predict the overall appeal of images with results closely corresponding to human ranking.
\end{abstract}
\vspace{-2mm}
%%%%%%%%% BODY TEXT
\section{Introduction}
\label{sec:intro}
With smartphones and digital cameras being ubiquitous, users now capture and share a large number of images every day. According to a recent study, more than $1.2$ trillion images were clicked in 2017 alone \cite{survey}. As a result, users often end up with a large collection of images with no good way to organize or navigate through them. Popular photo managers, such as Google photos and Flickr, let users organize images based on date, time and several other tags, such as places or events \cite{googlephotosurl,flickrurl}. This is often done using the meta-data embedded by the capturing devices in the photos, and does not require analyzing the actual content of the images.

\begin{figure}[tb]
    \centering
    \includegraphics[width=0.9\linewidth]{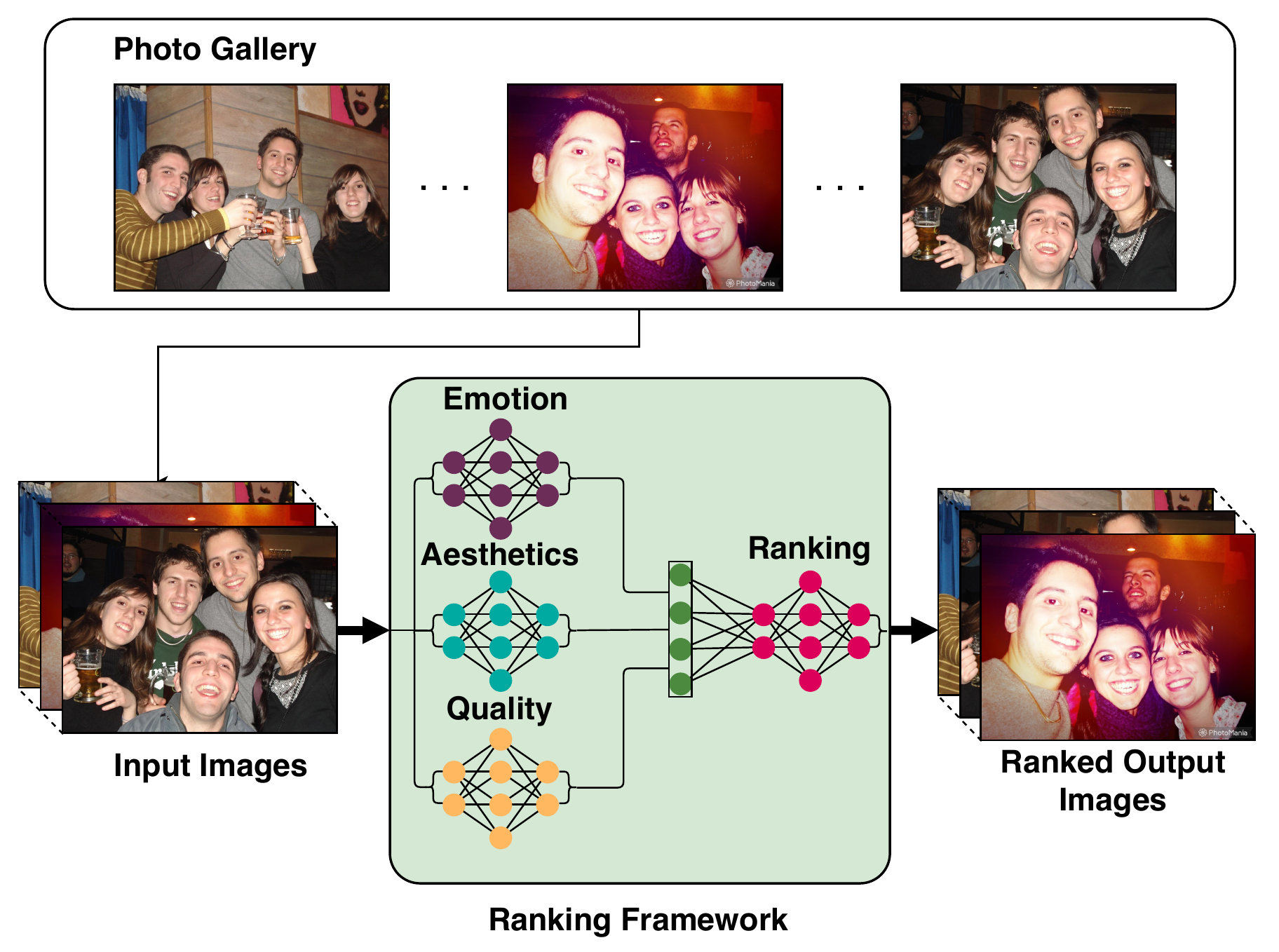}
     \vspace{-1em}
    \caption{Proposed ensemble network for ranking group images.}
     \vspace{-1.5em}
    \label{fig:overview}
\end{figure}

One common reason for accumulating a large number of photos is multiple clicks (not always very rapid) of the same scene with moderate changes in viewpoint, scale, illumination, color, pose and facial expressions (see Fig.~\ref{fig:overview}, Fig.~\ref{fig:database_sample}). In this paper, we address the problem of ranking images of the same scene with moderate changes in order to improve user experience of navigating through large photo libraries. In particular, we consider ranking group photos, where two or more people are photographed in the same scene with moderate changes in viewpoint, scale, illumination, color, pose and facial expressions. This is a challenging special case of the general image ranking problem because it demands fine-grained ranking of a set of images that may not have significant semantic differences.

In this paper, we propose a group photo ranking framework that combines \emph{group emotion}, \emph{image aesthetics}, and \emph{image quality} estimated from the image content. Our framework can be seen as a fusion of three channels of information pertaining to group emotion, aesthetics, and visual quality followed by a ranking network. Fig.~\ref{fig:overview} presents the overall idea of our proposed framework. First, we propose an end-to-end convolutional neural network (CNN) architecture, called the \emph{saliency-enforced CNN (sCNN)}, for accurately predicting group emotion in images. The main idea behind the proposed architecture is to force the network to put more weight on the salient regions of an image while estimating group emotion. In order to extract aesthetic information from images, we use a pretrained CNN model \cite{naest3} that yields an image aesthetics score.  A traditional no-reference image quality assessment method (the blind/reference-less image spatial quality evaluator (BRISQUE) \cite{BRISQUE}) is used to obtain a score for perceptual image quality. After we obtain the scores corresponding to group emotion, aesthetics and image quality for each set of group photos, we accumulate the scores in a single vector and feed to a ranking network. To this end, we also curated a new group photo database, called the \emph{ranked group photos} (rGroup) database, that contains $70$ sets of group photos (each set contains 3 images) taken during various events (both indoor and outdoor) with varying number of human subjects. Due to the small size of this database, end-to-end training was not possible. Hence, we separately train the group estimation and aesthetics CNN models on relevant databases, and use the data augmented version of our rGroup database for training and testing only the ranking network.

The contributions of this work are: (i) A new end-to-end CNN architecture (the sCNN) for predicting group emotion, which achieves state-of-the-art performance on benchmark database\footnote{Code link for sCNN: \texttt{https://github.com/deciphar/sCNN}}, (ii) A mulitchannel computational framework for ranking group photos based on group emotion, image aesthetics and image quality. (iii) A fully annotated database of group photos, and two new metrics for evaluating ranking performance.
%---------------------------------------------------------------------------------------------------------
\section{Related work}
\label{sec:related}
In this section, we review the related work on group emotion estimation and image ranking. \\

\vspace{-3mm}
\noindent\textbf{Group emotion.}
Group-level emotion estimation from images is an emerging topic of research. Most of its development can be credited to the recent release of large scale group emotion analysis databases, such as HAPPEI \cite{dhall2015automatic} and Group Affect 2.0 \cite{dhall2017individual}. The task of group-level emotion estimation is more complex than emotion estimation of individuals due to the dynamic nature of group size, variability in the individual expressions, context, and the subjectivity in the perception of overall group emotion. Dhall et al.~\cite{dhall2015automatic} proposed to estimate a group happiness intensity score using global and local features. Li and Roy \cite{li2016happiness} used a CNN to extract individual facial features and a recurrent neural network (RNN) to select the salient features to predict the group-level happiness. More recently, Rassadin et al. \cite{rassadin2017group} proposed to classify a group photo into positive, negative or neutral categories by first computing holistic features from the entire image using CNN, and then using them in an ensemble of random forests. Note that none of these models above involve end-to-end learning.\\

\vspace{-3mm}\noindent\textbf{Image ranking.}
The existing works on image ranking usually compared images based on aesthetics \cite{dhar2011high,nishiyama2011aesthetic,lu2014rapid,lu2015deep} and visual image quality \cite{kang2014convolutional}. Assuming that the modern cameras can capture high quality images very easily, several recent works propose to focus on aesthetic cues only \cite{lu2014rapid,lu2015deep,naest3}. 
In general, the papers mentioned above rely on low-level handcrafted features (hue, saturation, color) and photography rules, and traditional machine learning techniques, such as support vector regression (SVR). Several researchers \cite{lu2014rapid, lu2015deep, kang2014convolutional} pose ranking as a binary classification task with two labels: low or high aesthetics. With the introduction of large annotated aesthetics databases \cite{murray2012ava, naest3}, deep models are now being used to quantify aesthetics. Shu et al. \cite{naest3} developed a deep network for ranking closely related images based on aesthetics alone, and achieved state-of-the-art results.
%----------------------------------------------
\section{Database creation, human factors}
\label{sec:database}
\begin{figure}[tb]
    \centering
    \includegraphics[width=0.8\linewidth, trim={0cm 4.1cm 0cm 0cm}, clip=true]{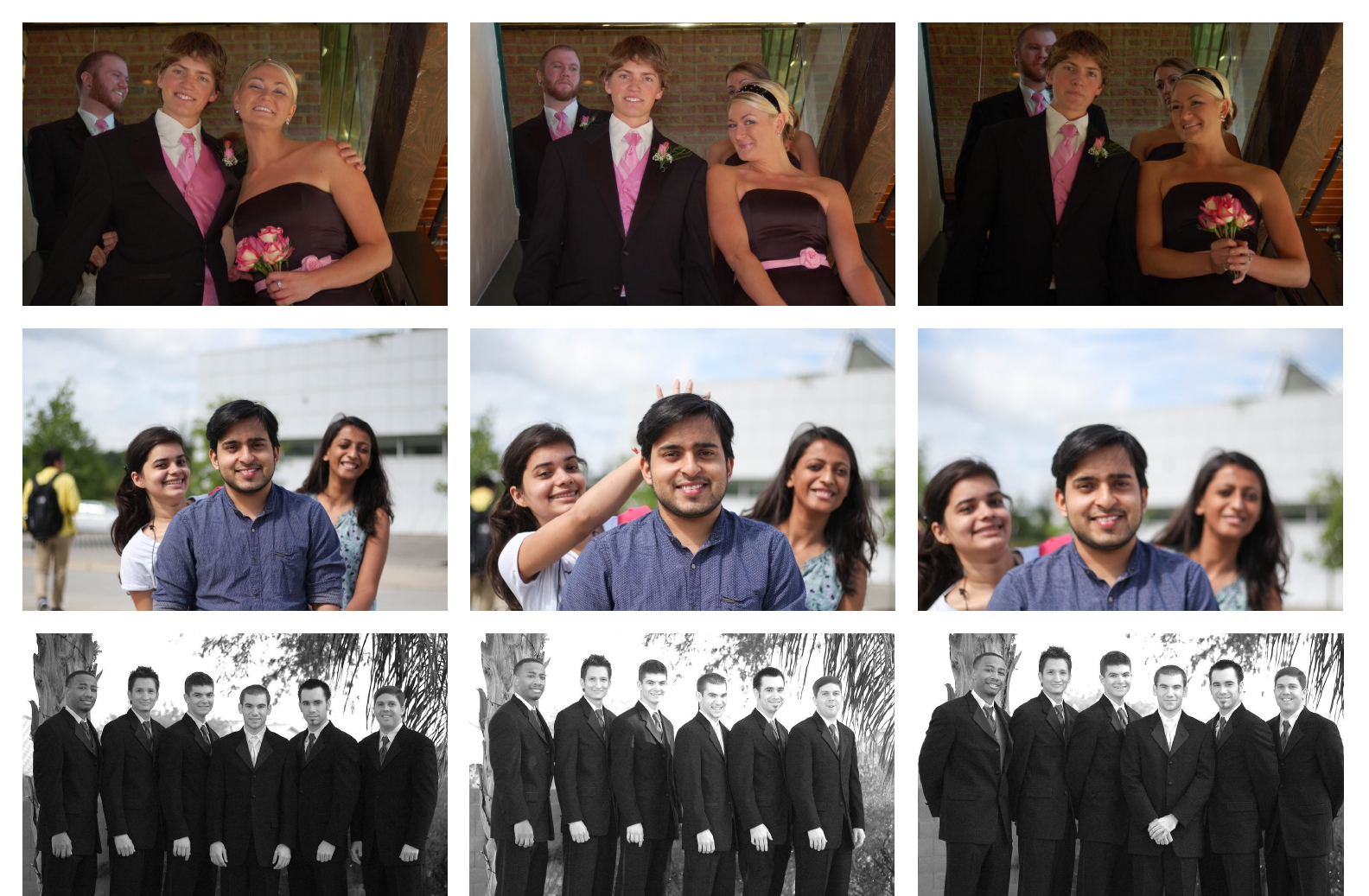}
    \vspace{-2mm}
    \caption{Sample images from our group photo database}
    % \vspace{-1.5em}
    \label{fig:database_sample}
\end{figure}

In order to better understand the factors humans take into account while choosing one group photo over another, we created a small database of group photos and collected manual annotation for subjective analysis. We created the rGroup database containing $70$ sets of group photos (each set having $3$ images) that were captured within a short time span. The images exhibit high variability in terms of context (indoor/outdoor, day/night, formal/casual events), scale, illumination, pose, viewpoint and number of subjects. Fig.~\ref{fig:database_sample} shows sample images from our rGroup database.
A list of eight visual features that are likely to be important for ranking were selected: \emph{group happiness, occlusion, motion blur, group pose, image quality, face size, face pose}, and \emph{eyes (closed or not)}. We then asked annotators to rank the images within each set according to their preference. Additionally, they were also asked to select the most relevant feature from the above list which has influenced their decision. We created a website for automatically collecting the annotations from as many annotators as possible. After collecting the annotations, incorrect annotations (e.g.~same rank assigned to all images) were manually discarded. Finally, each image received at least $5$ valid annotations. Final ground truth ranks were assigned based on majority voting, i.e., given a set, the image with the highest votes for rank 1 was labeled as  rank 1. In the cases of ties, the authors themselves acted as additional annotators to break the ties. Visual feature annotations were available for $45$ sets only as this was an optional question to the annotators.

\begin{figure}[tb]
    \centering
    \includegraphics[width=0.8\linewidth]{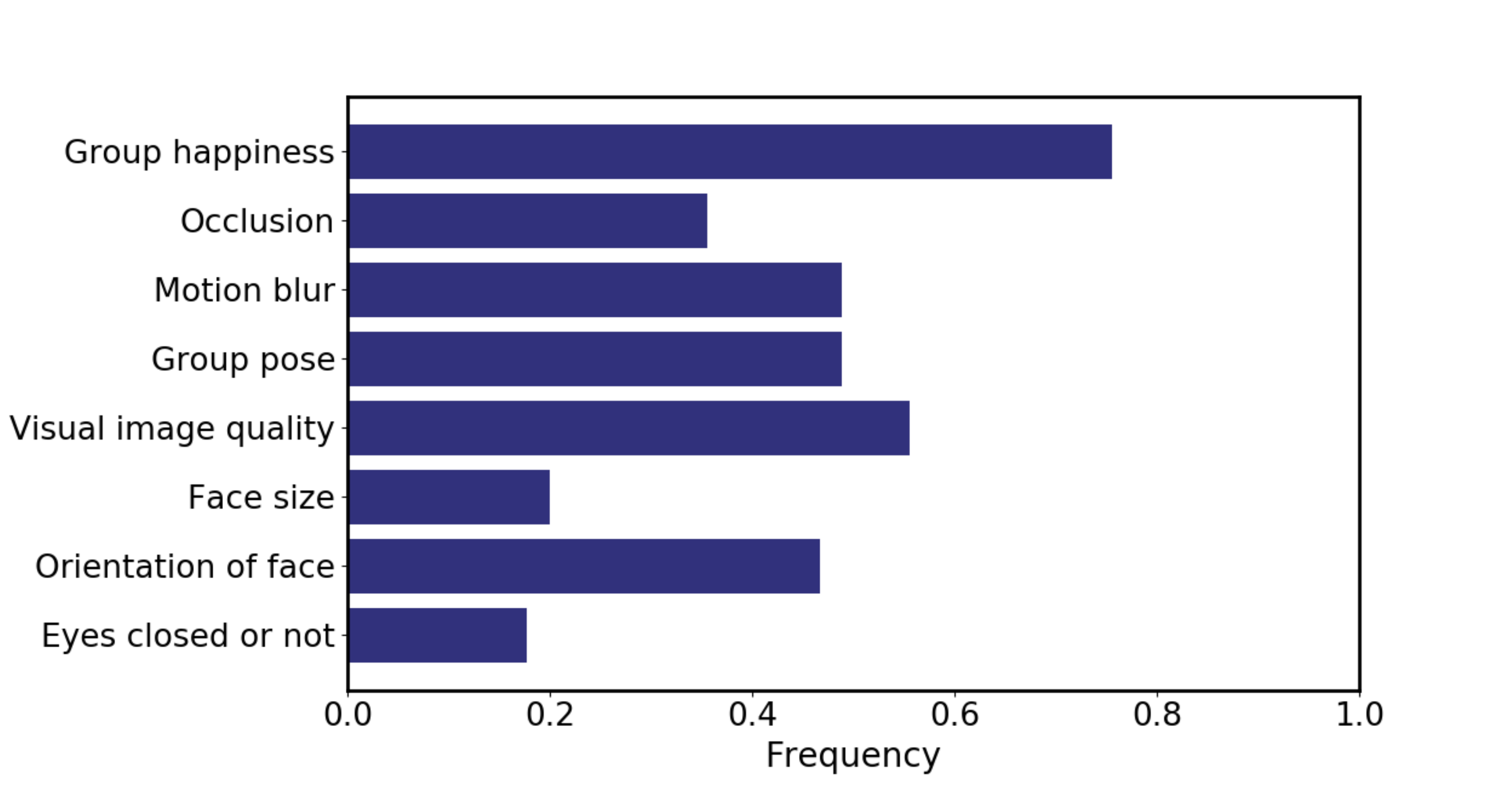}
     \vspace{-1em}
    \caption{Summary of annotator responses on preferred feature for ranking group photos.}
    % \vspace{-1em}
    \label{fig:prelim_results}
\end{figure}

Fig.~\ref{fig:prelim_results} summarizes the annotators' responses on the preferred  features for their ranking decision. Clearly, group happiness appears to be a frequently chosen feature. Among the rest, image quality, group pose and motion blur are three most relevant features that affects human's perception. Since motion blur is also a feature related to visual image quality, we observe that \emph{group happiness} and \emph{image quality} are the two most important factors while ranking group photos. This motivates the design of our model described in the following section.

%--------------------------------------
\section{Proposed image ranking framework}
\label{sec: proposed}
 The results in Section \ref{sec:database} informed that the two most frequently used visual cues for group photo ranking are group happiness and visual quality. Based on this observation, we now develop a ranking framework that estimates group emotion, aesthetics, image quality, and combine them for decision making. Fig.~\ref{fig:overview} presents an overview of the proposed framework. Below, we describe each part of the framework in detail.
\begin{figure*}
    \begin{center}
    \includegraphics[width=0.94\textwidth,height=0.5\textwidth,keepaspectratio]{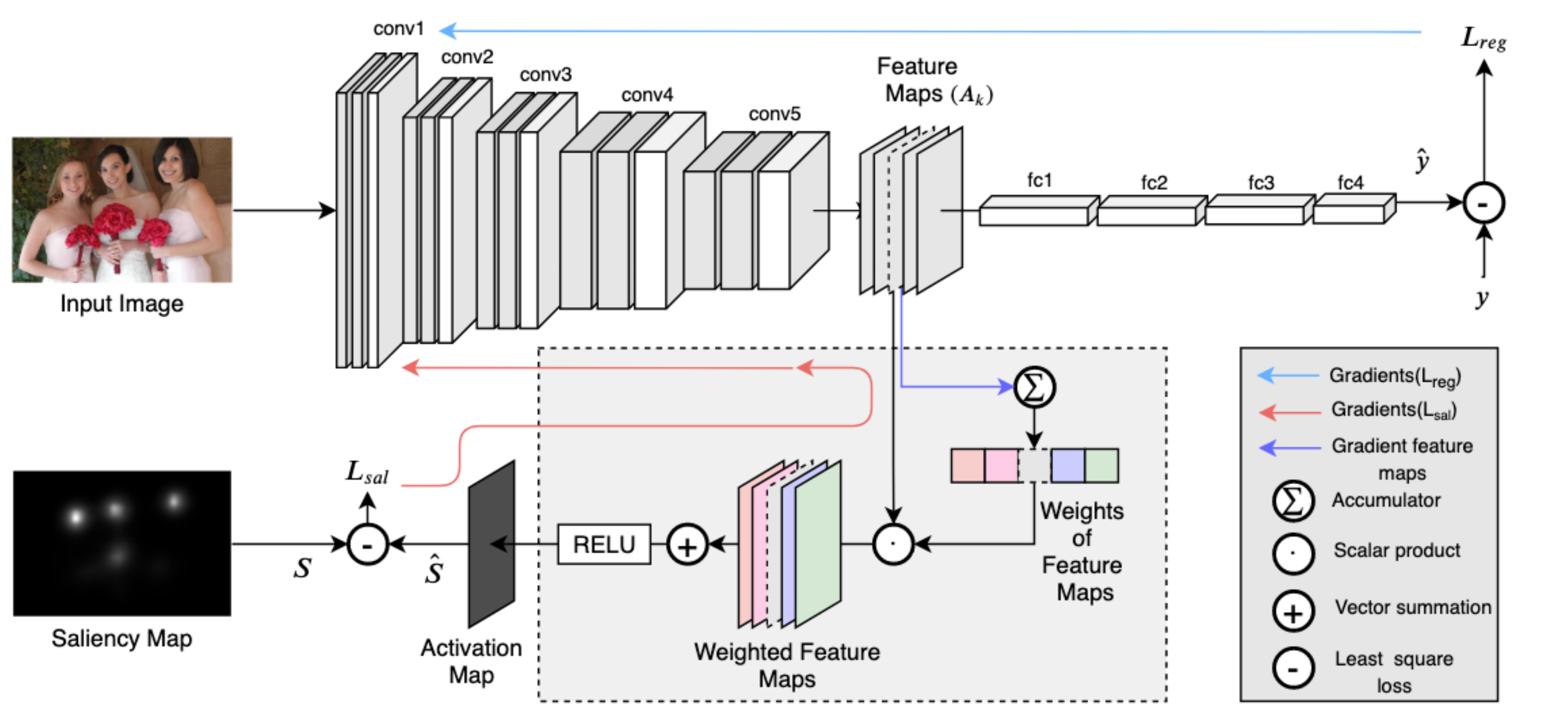}
    \end{center}
    % \vspace{-1em}
    \caption{Proposed architecture (sCNN) for group emotion estimation.}
    % \vspace{-1.5em}
    \label{fig:vgg-sal}
\end{figure*}
\vspace{-2mm}
\subsubsection*{4.1. Proposed network for estimating group happiness}
We propose a CNN-based architecture (Fig.~\ref{fig:vgg-sal}), called the sCNN, to estimate a group happiness score from an image. The architecture forces the network to concentrate on the salient regions of an input image. This information is provided to the network as an additional input in form of a saliency map.

Our network takes $N$ training images $X_1, \cdots X_N$ and their corresponding saliency maps $S_1, \cdots S_N$ as inputs. The saliency maps are precomputed using a state-of-the-art model \cite{sam}. For a given test image, this network outputs an estimated group happiness score $\hat{y}$ and its attention map $\hat{S}$. 
In order to obtain $\hat{S}$, we first compute the gradients of all the feature maps $A^{(k)}$ with respect to the final prediction score $\hat{y}$ i.e., $\frac{d\hat{y}}{dA^{(k)}}$. Then we average-pool the gradients of all the neurons within a feature map to compute the relative weights $w^{(k)}$ of that activation map. 
We then compute a weighted average of all the activation maps followed by a non-linear activation. 
\begin{align*}
     \hat{S} &= \mathrm{ReLU}\Big(\sum_k w^{(k)} A^{(k)}\Big)
\end{align*}
The regression loss $L_{reg}$ and the saliency loss $L_{sal}$ are computed as
\begin{align*}
    L_{reg} &= \frac{1}{N}\sum_{i=1}^N\left\|\hat{y}_i-y_i\right\|_2^2\\ %\label{eq:reg_loss}\\
    L_{sal} &=  \frac{1}{N}\sum_{i=1}^N\left\|\hat{S}_i-S_i\right\|_2^2 %\label{eq:sal_loss}
\end{align*}
While $L_{reg}$ helps in optimizing the network on the group happiness estimation task, $L_{sal}$ complements it by forcing the network to attend to the salient regions. The learnable parameters $\Phi:\{\Phi_c,\Phi_d\}$ of the networks are updated as follows:
\begin{align}
    \Phi_c &\gets \Phi_c + \eta(\frac{dL_{reg}}{d\Phi_c}+\lambda_1\frac{dL_{sal}}{d\Phi_c}) \label{eq:update_c} \\
    \Phi_d &\gets \Phi_d + \eta(\frac{dL_{reg}}{d\Phi_d} \label{eq:update_d})
\end{align}
Note that $\Phi_c$ (for the convolution layers) gets updated with both the losses while $\Phi_d$ (for the dense layers) uses only $L_{reg}$ (see Fig.~\ref{fig:vgg-sal}). The relative weight of the losses in (1) is adjusted using $\lambda_1$, and $\eta$ is the learning rate.
%------------
\vspace{-2mm}
\subsubsection*{4.2. Quantifying aesthetics}
Recently, Shu et al.~\cite{naest3} have shown that aesthetics is a key factor in fine-grained image ranking. We adopt their CNN architecture trained as a siamese network for quantifying aesthetics. Thus our aesthetics network has the following loss function:
\begin{align}
    L &= L_{reg}  + \frac{\lambda_2}{2N}\sum_{i,j} \mathrm{max}(0, \alpha -\delta(y_i, y_j)(\hat{y_i}-\hat{y_j})) \label{eq:ast_rank_loss}
\end{align}
where $\alpha$ is a specified margin parameter, and 
 $\delta(y_i,y_j)$ is defined as 
\begin{equation}
\delta(y_i,y_j) =
            \begin{cases}
                1 & y_i \ge y_j \\
                -1 & y_i < y_j  
            \end{cases} 
\end{equation}
\vspace{-2mm}
\subsubsection*{4.3. Estimating image quality} 
For this subtask, we use a traditional no-reference image quality metric, called BRISQUE \cite{BRISQUE}. It uses natural scene statistics of locally normalized luminance coefficients to quantify the loss of the naturalness in a distorted image. After extracting various statistical measures from an image, BRISQUE trains an SVR to predict the quality score.
\vspace{-2mm}
\subsubsection*{4.4. Multichannel fusion and ranking}
To combine the information from the three channels (group happiness, aesthetics, image quality), each channel score $\kappa_c$ is first normalized to lie between 0 and 1, and then transformed to a 2D vector $\kappa_c = [\kappa_c,\kappa_c^2]$. We concatenate the vectors from all the three channels to produce a 6 dimensional vector $\kappa$ for each image, which is fed to the ranking network. We use two ranking methods: (i) a rank support vector machine (rankSVM) \cite{ranksvm}, and (ii) a shallow neural network (rankNet) with the ranking loss. 
For the rankNet, we use a ranking loss similar to the second term in (3), i.e.,
\begin{equation*}
\label{eq:rankloss}
%    \begin{split}
        L_{rank} = \mathrm{max}(0, \alpha -\delta(y_i, y_j)(\hat{y_i}-\hat{y_j}))
\end{equation*}
\newline where $\delta$ is defined as before. 
In contrast to the work of Shu et al.~\cite{naest3}, we do not add any regression loss. This is because our dataset contains relative rank labels only. 
%-----------------
%
\section{Performance evaluation}
\label{sec:performance}
In this section, we first present the results of the proposed sCNN architecture for group happiness estimation. Subsequently, we present results on group photo ranking.
\subsection{Results on group happiness estimation}
\textbf{\textit{Database.}} We use the popular HAPPEI database \cite{dhall2012fin} for evaluating the performance of our architecture. It contains around 3,000 images with high variability in terms of illumination and background. Every image is labeled with one of the 6 discrete group happiness intensity (0 to 5). The labels are available both for individuals and the group.\\  

\vspace{-3mm}\noindent\textbf{\textit{Implementation details.}}
We precomputed the saliency maps for all the training images. 
We first train the sCNN network \emph{only} for the group happiness prediction task i.e., without the saliency branch. After that, we train the network including the saliency branch, and update the weights on both the branches. Our network is trained with $\lambda = 1e-3$, $\eta = 1e-4$, batch size of 5 and stochastic gradient decent (SGD) optimizer.\\

\vspace{-3mm}\noindent\textbf{\textit{Results.}} The performance is evaluated in terms of mean absolute error (MAE). The results are presented in Table \ref{tab:group_emotion}. The proposed architecture outperfornificant margin, and produces comparable result with the state-of-the-art \cite{dhall2012fin}. The results also shows that the saliency information helps to improve the overall accuracy significantly.
\subsection{Results on group photo ranking}
\textbf{\textit{Database.}} The results are evaluated using the \emph{rGroup} database. The details of the database are presented in Section \ref{sec:database}. \\

\vspace{-3mm}\noindent\textbf{\textit{Proposed evaluation metrics.}} To evaluate the ranking performances, we propose two new evaluation metrics: (i) Best image match (BIM) and (ii) Percentage of swapped pairs (PSP).\\
BIM (expressed in \%) is defined as: 
\begin{equation*}
    BIM = \frac{\text{total number of true positive pairs}}{\text{total numbers of sets}}
\end{equation*}
where a true positive pair is defined as the pair when the predicted highest rank image matches the ground truth. \\
The PSP metric (expressed in \%) is defined as: 
\begin{equation*}
    PSP = \frac{\sum_i^N\text{total number of swapped pairs in the $i^{th}$ set}}{\sum_i^N\text{total number of possible pairs in $i^{th}$ set }}
\end{equation*}
where a swapped pair is defined as follows: Let $R(I_i^s), R(I_j^s)$ denote the true ranks of the $i^{th}$ and $j^{th}$ images within a set $s$, and $\hat{R}(I_i^s), \hat{R}(I_j^s)$ be their predicted ranks. If the relative ordering of $R(I_i^s), R(I_j^s)$ does not match the relative order of $\hat{R}(I_i^s), \hat{R}(I_j^s)$ then this is considered as a swapped pair.\\

\vspace{-3mm}\noindent\textbf{\textit{Implementation details}.} The details of the group happiness estimation channel is already presented in Section 5.1. For quantifying the aesthetics, we used a CNN network pretrained over the Aesthetics and Attributes database \cite{naest3} in a siamese fashion. AADB is a large database with over 10k images, where each image is annotated with aesthetic quality ratings and aesthetics attributes. To choose among the various image quality estimators available, we ran multiple experiments and based on the results we select BRISQUE \cite{BRISQUE} as it yielded the best result on the validation set. The BRISQUE scores are unbounded. We normalized them to lie between 0 to 1.
%---------------
\begin{table}[tb]\centering
\caption{Group happiness estimation results on HAPPEI database.}
\label{tab:group_emotion}
 \renewcommand*{\arraystretch}{1.2}
\vspace{1mm}
\begin{tabular}{l  c}
\toprule
\textbf{Method} & \textbf{MAE} $\downarrow$ \\
\midrule
Mean emotion \cite{dhall2012fin} & 0.57 \\
Dhall et al. \cite{dhall2012fin} & \bf 0.38\\
Proposed without saliency & 0.42 \\ 
Proposed sCNN & \bf 0.39 \\ 
\bottomrule
\end{tabular}
\end{table}
%------------
%---------
\begin{table}[tb]
\label{tab:rank_per}
 \renewcommand*{\arraystretch}{1.2}
    \caption{Group photo ranking performance on rGroup database.}
    \vspace{1mm}
    \centering
    \begin{tabular}{l  c c c}
    \toprule
    \bf{Method} & \bf BIM $\uparrow$ & \bf PSP $\downarrow$ & \bf Corr ($\rho$)$\uparrow$\\
    \hline \hline
    Avg. human performance & 74.00 & 7.95 & 0.93\\
    \hline
    \multicolumn{4}{c}{\emph{Individual channel}}\\
    \hline
    Group happiness (sCNN) & 27.14 & 39.70 & 0.21 \\
    
    Aesthetics \cite{naest3} & 37.10 & 27.80 & 0.52\\
    
    Image quality &  47.14 & 22.04 & 0.65 \\
    \hline
    \multicolumn{4}{c}{\emph{All channels}}\\
    \hline
    Mean pooling & 40.00 & 22.61 & 0.63 \\
    Max pooling & 41.40 & 27.85  & 0.52 \\
    rankSVM & 48.60  & 21.85 & \bf 0.69 \\
    rankNet & \bf 52.38 & \bf 18.00 & \bf 0.69\\
         \bottomrule
    \end{tabular}
\end{table}
%
%---------
Two ranking approaches were used: rankSVM and rankNet. In both cases, we use a 5-fold cross-validation scheme. The proposed rankNet consists of a pair of two fully connected neural networks, where each network has the following four layers - linear ($3\times 3$), ReLU ($3\times 3$), linear ($3\times 1$) and ReLU ($1\times1$). The two networks are combined and trained as a siamese network using the loss function given by Eq.~\eqref{eq:rankloss}. %We set $m = 2$ (margin parameter).
In each experiment, we trained the network for 100 epochs with a learning rate $=1e-5$, batch size $=5$.\\

\vspace{-3mm}\noindent \textbf{\textit{Results.}} Table 2 presents all results on group photo ranking on the rGroup database. Overall, rankNet shows the best performance in terms of the BIM and PSP metrics, while rankSVM performs the best in terms of correlation ($\rho$). The superior performance of the rankNet can be largely attributed to the non-linearity of the rankNet, while rankSVM being a linear function.  The results are compared with human performance and random chance. We also investigated the performance of each channel for the overall ranking task using three evaluation metrics, and observed that the image quality channel performs the best.
Simple feature pooling techniques were also used to predict the ranks, i.e., without rankSVM or rankNet. For these experiments, either the max of the 3 values (max pooling) or the average of the 3 values (mean pooling) were used as the final score. 
\vspace{-2mm}
\section{Conclusion}
We proposed a computational framework for ranking group photos with moderate variations in illumination, scale, viewpoints and even group size. Our framework is an ensemble of three channels that extract emotion, image quality and aesthetics from images. We proposed a new architecture for group happiness estimation, and the overall framework for image ranking. We also created a labeled database, and proposed two metrics for evaluating ranking performance. The proposed ranking framework achieves high correlation with human perception, and outperforms existing works involving a single channel. Future work will be directed towards building an end-to-end network with evaluations on a larger database. 
%
% -------------------------------------------------------------------------
\newpage
\balance
\bibliographystyle{IEEEbib}
\bibliography{egbib,strings}
\end{document}